%% file: main.tex
\documentclass{article}
\PassOptionsToPackage{numbers, compress}{natbib}

\usepackage[preprint]{neurips_2025}

\usepackage[utf8]{inputenc} 
\usepackage[T1]{fontenc}    
\usepackage{hyperref}       
\usepackage{url}            
\usepackage{booktabs}       
\usepackage{amsfonts}       
\usepackage{nicefrac}       
\usepackage{microtype}      
\usepackage{xcolor}         
\usepackage{multirow}
\usepackage{graphicx}

\usepackage{tcolorbox}
\usepackage{fancyvrb}
\usepackage{fvextra}

\DefineVerbatimEnvironment{VerbatimWrap}{Verbatim}{
breaklines=true, 
breakindent=0pt, 
breaksymbol={},
fontsize=\small,
}
\newcommand{\hfmodel}[1]{\texttt{#1}}

\title{An Open and Reproducible Deep Research Agent\\for Long-Form Question Answering}

\author{
  Ikuya Yamada$^{1,2}$ \And
  Wataru Ikeda$^{2}$ \And
  Ko Yoshida$^{2}$ \And
  Mengyu Ye$^{2}$ \AND
  Hinata Sugimoto$^2$ \And
  Masatoshi Suzuki$^{1,2}$ \And
  Hisanori Ozaki$^3$ \And
  Jun Suzuki$^{2,4,5}$
  \AND
  \normalfont
  $^{1}$Studio Ousia\\
  \texttt{\small{\{ikuya,m.suzuki\}@ousia.jp}} \\[0.5em]
  $^{2}$Tohoku University\\
  \texttt{\small{\{ikeda.wataru.q5,yoshida.kou.p3,ye.mengyu.s1,sugimoto.hinata.q2\}@dc.tohoku.ac.jp}}\\
  \texttt{\small{jun.suzuki@tohoku.ac.jp}} \\[0.5em]
  $^{3}$DENTSU SOKEN INC.\\
  \texttt{\small{ozaki.hisanori@dentsusoken.com}}\\[0.5em]
  $^{4}$RIKEN,
 $^{5}$NII LLMC
}

\begin{document}
\maketitle
\begin{abstract}
We present an open deep research system for long-form question answering, selected as a winning system in the text-to-text track of the MMU-RAG competition at NeurIPS 2025.
The system combines an open-source large language model (LLM) with an open web search API to perform iterative retrieval, reasoning, and synthesis in real-world open-domain settings.
To enhance reasoning quality, we apply preference tuning based on LLM-as-a-judge feedback that evaluates multiple aspects, including clarity, insightfulness, and factuality.
Our experimental results show that the proposed method consistently improves answer quality across all three aspects.
Our source code is publicly available at \url{https://github.com/efficient-deep-research/efficient-deep-research}.
\end{abstract}

\section{Introduction}

The text-to-text track of the MMU-RAG competition\footnote{\url{https://agi-lti.github.io/MMU-RAGent/}} at NeurIPS 2025 focuses on building retrieval-augmented generation (RAG) systems that generate long-form answers under real-world conditions.
A natural approach to developing such RAG systems is to adopt deep research, a tool-augmented LLM capable of performing retrieval, reasoning, and synthesis to address a broad range of questions.

Long-form question answering based on deep research has been implemented in commercial systems \citep{openai2025deepresearch,geminideepresearch}, but these systems remain proprietary, making them difficult to study or reproduce.
Motivated by these limitations, recent studies have attempted to reproduce similar capabilities~\citep{li2025search,wang2025chainofretrievalaugmentedgeneration,sun2025simpledeepsearcher,jin2025searchr,song2025r1,zheng2025deepresearcher,li2025webthinker,sun2025zerosearch,gao2025beyond,nguyen2025sfr}.

However, most of these studies have focused on short-form question answering, where answers can be easily verified against gold-standard references, while only a limited number of recent works address long-form question answering based on deep research~\citep{li2025webthinker,nguyen2025sfr,shao2025dr}. 

In this report, we present our deep research system submitted to the text-to-text track of the MMU-RAG competition.
The system is designed to effectively address long-form question answering and is built using an open-source LLM and a search API developed on an open web corpus, promoting reproducibility.
Our system integrates high-quality synthetic data generation, preference tuning, and an improved search component, yielding substantial gains across evaluation metrics.

We adopt the metrics proposed by~\citet{coelho2025deepresearchgym}—\texttt{clarity}, \texttt{insightfulness}, and a factuality metric based on KPR~\citep{qi-etal-2024-long2rag}, which we refer to as \texttt{factuality} throughout this paper,\footnote{See Section~\ref{subsec:gen_pref_pair} for precise definitions.} and train an LLM using Direct Preference Optimization (DPO)~\citep{NEURIPS2023_a85b405e}.
Following prior work~\citep{li2025webthinker,nguyen2025sfr}, we use an LLM to evaluate preferences along these dimensions.

We also extend the search API with a reranking and summarization module. 
After retrieving documents, the module reranks them and produces concise summaries of the top reranked documents, enabling the generator to produce more accurate and better-supported answers.
The system also produces inline citations by prompting the model to link statements to the corresponding documents.

Experiments using an LLM-as-a-judge setup show consistent improvements across all three metrics.
These findings indicate that preference-based tuning, combined with enhanced retrieval and summarization, leads to measurable performance gains for long-form deep-research tasks.
Furthermore, the system received the Best Static Evaluation award in the open-source category of the text-to-text track.

\begin{figure}[t]
    \centering
    \includegraphics[width=\linewidth]{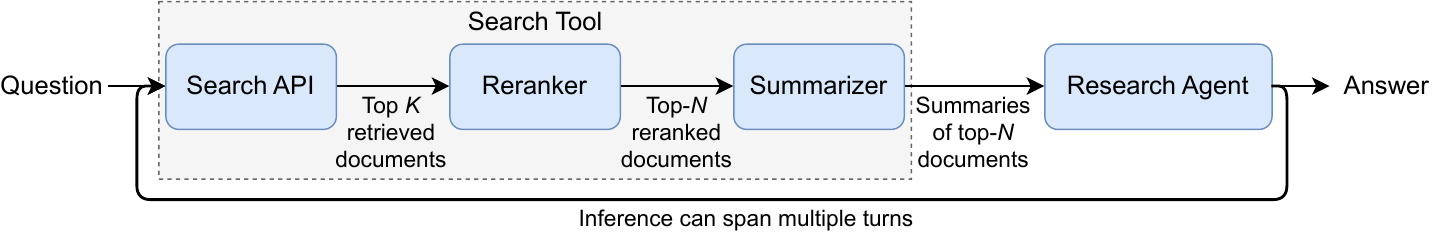}
    \caption{Architecture of our deep research system.}
    \label{fig:architecture}
\end{figure}

\section{Architecture}

Figure~\ref{fig:architecture} illustrates the architecture of our system.
The system consists of a search tool and a research agent.
The search tool takes a query and returns summaries based on the retrieved documents.
The research agent manages the reasoning flow by iteratively invoking the search tool to gather and synthesize information into a coherent final answer.

\subsection{Search Tool}

The search tool is a pipeline comprising a search API, a reranker, and a summarizer.

\paragraph{Search API}

We use the open search API \citep{coelho2025deepresearchgym}, which is built on ClueWeb22 \citep{10.1145/3477495.3536321}, a large collection of billions of web documents.\footnote{We use the ClueWeb22-A category covering two billion pages.}
For each query, the top-$K$ documents are retrieved from the API and passed to the reranker.
Here, the query can be either the original question or one generated by our research agent.

\paragraph{Reranker}

The reranker reorders the top-$K$ documents returned by the search API and selects the most relevant ones.
We use the state-of-the-art model \hfmodel{Qwen3-Reranker-0.6B}\footnote{\url{https://huggingface.co/Qwen/Qwen3-Reranker-0.6B}}
 \citep{zhang2025qwen3}, with the prompt described in Appendix~\ref{sec:prompts}.
The top-$N$ reranked documents are then passed to the summarizer.

\paragraph{Summarizer}

The summarizer extracts information relevant to the given query and previous reasoning steps, generating a concise summary of the retrieved web documents.
We use \hfmodel{Qwen3-Next-80B-A3B-Thinking}\footnote{\url{https://huggingface.co/Qwen/Qwen3-Next-80B-A3B-Thinking}} \citep{yang2025qwen3} with the prompt described in Appendix~\ref{sec:prompts}.
Using a simple yet effective prompting technique, we instruct the model to include citation markers in the summary, each corresponding to its source document.

\subsection{Research Agent}

The research agent manages the overall reasoning process of the system.
It takes the question and the previously generated summaries as input, determines whether additional searches are needed through reasoning, and issues a new query when necessary.
The agent iteratively calls the search tool described above and integrates the gathered information into a coherent final answer.
Before starting the reasoning process, we invoke the search tool using the input question as the query and include the summarized results in the research agent’s prompt.
We employ \hfmodel{Qwen3-Next-80B-A3B-Thinking}, fine-tuned as described in the next section, with the prompt provided in Appendix~\ref{sec:prompts}.

\section{Training of Research Agent}

We provide details on the creation of the synthetic data and the DPO training of our system.

\subsection{Data Generation}
\label{subsec:gen_pref_pair}

The construction of preference pairs consists of four steps: (1) collecting questions for training, (2) generating answers to these questions, (3) performing LLM-based preference evaluation, and (4) constructing preference pairs based on the evaluation results.

\paragraph{Metrics}
We use the \texttt{clarity} and \texttt{insightfulness} metrics defined by~\citet{coelho2025deepresearchgym}, and we use the KPR metric~\citep{qi-etal-2024-long2rag}, which we refer to as \texttt{factuality}. Specifically, \texttt{clarity} reflects logical coherence and linguistic fluency; \texttt{insightfulness} captures analytical nuance and depth of reasoning; and \texttt{factuality} measures consistency with the underlying facts.\footnote{See Appendix~\ref{sec:prompts} for the prompts used to assess each metric.}

\paragraph{Question Collection}
We collect 1,000 training questions from multiple datasets: 500 from Researchy Questions~\citep{rosset2024researchyquestions}, 200 from Natural Questions~\citep{kwiatkowski-etal-2019-natural}, and 300 from the official validation set provided by the competition organizers, resulting in a balanced dataset that includes both factoid and non-factoid questions and reflects the diversity observed in real-world usage and in the competition’s evaluation.
Researchy Questions consists of non-factoid questions collected from a commercial search engine, accompanied by URLs clicked by users for each query. We collect web documents from these URLs to extract key points used in the \texttt{factuality} evaluation. Natural Questions primarily consists of factoid questions sampled from a commercial search engine. The competition’s official validation set provides questions designed to match those used in the final evaluation.

\paragraph{Answer Generation}
We generate answers for the questions using the research agent based on an untuned LLM.
For each question, we perform 20 sampling runs.
The hyperparameters used during sampling are described in Appendix~\ref{sec:generation_hparams}.

\paragraph{Preference Score Evaluation}

We use OpenAI's \hfmodel{o3-mini} to assign preference scores to the generated answers.
Specifically, we evaluate all three metrics on Researchy Questions, and only \texttt{clarity} and \texttt{insightfulness} for Natural Questions and the validation set, since the user-clicked URLs required to assess \texttt{factuality} are not available for these datasets.
The final score is obtained by sum up the normalized score of each metrics.

\paragraph{Preference Pair Construction}

We construct the DPO data using the preference scores and the number of search queries used to generate each answer. The answer with the highest preference score is selected as the \textit{chosen} response, and the one with the lowest score as the \textit{rejected} response. When answers share the same preference score, we select the answer with more search queries as the \textit{chosen} response, and vice versa.

We then filter out answers containing formatting errors, such as incorrect citation of the source. Finally, to retain only preference pairs with a meaningful score gap, we set a minimum threshold~$\theta$ for the score difference between the chosen and rejected responses and include only the pairs that satisfy this threshold.

\subsection{Training}

We perform DPO training using LoRA\footnote{For LoRA, we use $\alpha = 16$, $\mathrm{rank} = 16$, and for DPO, we use $\beta = 0.5$.} on the synthetic dataset, without extensive hyperparameter tuning. The model is optimized on both the reasoning chain and the final answer. We mask out the tokens corresponding to the responses returned by the search tool during training, as they are not generated by the research agent.

We construct three preference datasets by setting the minimum score-gap threshold to $\theta \in \{0.3, 0.5, 0.7\}$, yielding datasets with 983, 828, and 341 pairs. We train the model for one epoch on each dataset, resulting in $56$, $47$, and $20$ training steps. Details of the datasets are provided in Appendix~\ref{sec:preference_datasets}.

\section{Evaluation}

We verify the effectiveness of the our approach based on LLM-as-judge.

\subsection{Evaluation Setup}

We sample 100 queries from Researchy Questions that are not included in the training set and conduct LLM-based evaluation on each. 
Specifically, for each question in the evaluation data, we generate answers using one sampling run from the model before tuning (\textsc{Vanilla}) and from models tuned with different minimum score-gap thresholds $\theta$.
Because all tuned models consistently outperform \textsc{Vanilla} with only marginal differences among them, we hereafter focus on the model tuned with $\theta = 0.3$, referred to as \textsc{Ours}\footnote{See Appendix~\ref{sec:preference_datasets} for comparisons of evaluation results across tuned models with different threshold $\theta$.}\footnote{We selected the model tuned with $\theta = 0.3$ for competition submission.}. 
The hyperparameters for answer generation and the evaluation procedure follow those used for constructing the preference pairs in Section~\ref{subsec:gen_pref_pair}.
We also measure \texttt{search\_count} (the number of searches issued by the system) and \texttt{citation\_error\_rate} (the proportion of final answers with incorrectly formatted citation markers among the 100 samples) to analyze system behavior beyond answer quality.

\subsection{Results}

Table~\ref{tab:eval_results} shows our evaluation results. Our tuned model outperforms \textsc{Vanilla} across all LLM-based evaluation metrics, demonstrating the effectiveness of our training. The gains are particularly prominent for \texttt{clarity} ($+1.47$) and \texttt{insightfulness} ($+0.98$), while \texttt{factuality} shows a marginal improvement ($+0.41$). 
We believe this is partly because training for \texttt{factuality} is more difficult than for the other two metrics, as it relies on noisy user-click–based training signals and is available only for a subset of the training data.
The tuned model also performs more searches on average ($+0.17$), while exhibiting a slight increase in \texttt{citation\_error\_rate} ($+0.03$).

\input{eval_retults}

\section{Conclusion}

In this report, we presented an open and reproducible deep research system for long-form question answering, which was a winning system in the text-to-text track of the MMU-RAG competition at NeurIPS 2025.
The system is built by combining high-quality synthetic data creation, DPO training, and an improved search component.
Our system effectively addresses real-world question-answering tasks by integrating retrieval, reasoning, and synthesis during inference.
LLM-based evaluations confirm the effectiveness of our training, showing noticeable improvements across all metrics.

\section*{Acknowledgments}
This work was supported by the “R\&D Hub Aimed at Ensuring Transparency and Reliability of Generative AI Models” project of the Ministry of Education, Culture, Sports, Science and Technology, and JST Moonshot R\&D Grant Number JPMJMS2011-35 (fundamental research).

\bibliographystyle{plainnat}
\bibliography{references}

\newpage
\appendix

\section{Prompts}
\label{sec:prompts}
The prompts for the reranker and the research agent are shown in Figures~\ref{fig:prompt-reranker} and \ref{fig:prompt-research-agent}, respectively.
We use the same prompt as \hfmodel{Qwen3-Reranker-0.6B} \citep{zhang2025qwen3} for the reranker.

The prompts used for the summarizer are shown in Figures~\ref{fig:prompt-summarizer-initial} and \ref{fig:prompt-summarizer}.
We use different prompts for summarizing the search results included in the research agent’s initial prompt (Figure~\ref{fig:prompt-summarizer-initial}) and for summarizing the results of searches invoked during the agent’s reasoning process (Figure~\ref{fig:prompt-summarizer}).

We evaluate \texttt{factuality}, \texttt{clarity}, and \texttt{insightfulness} using the prompts shown in Figures~\ref{fig:prompt-factuality}, \ref{fig:prompt-clarity}, and \ref{fig:prompt-insightfulness}, respectively. For \texttt{factuality} evaluation, we sample 10 key points for each question.

\section{Details of Preference Datasets}
\label{sec:preference_datasets}
Table~\ref{tab:gap_comparison} shows the statistics of the three preference datasets constructed with different minimum score-gap thresholds $\theta \in \{0.3, 0.5, 0.7\}$.
Table~\ref{tab:eval_results_diff_theta} shows the evaluation results of models trained on each of these datasets.

\input{preference_data}
\input{eval_results_diff_theta}

\section{Hyperparameters for Answer Generation}
\label{sec:generation_hparams}
Table~\ref{tab:generation_hparams} shows the hyperparameters used for answer generation.
We use the same settings for both preference pair construction and evaluation.
To improve answer quality, we modify only the Search API top-$K$ to 300 in the competition evaluation.

\input{generation_hparams}

\begin{figure}
\begin{tcolorbox}[title=Reranker Prompt]
\begin{VerbatimWrap}
System prompt:
Judge whether the Document meets the requirements based on the Query and the Instruct provided. Note that the answer can only be "yes" or "no".

User prompt:
<Instruct>: Given a web search query, retrieve relevant passages that answer the query
<Query>: {query}
<Document>: {document}
\end{VerbatimWrap}
\end{tcolorbox}
\caption{The prompt used for the reranker. The placeholders \texttt{\{query\}} and \texttt{\{document\}} are replaced with the input query and document, respectively.}
\label{fig:prompt-reranker}
\end{figure}

\begin{figure}
\begin{tcolorbox}[title=Summarizer Prompt for Initial Search Results]
\begin{VerbatimWrap}
**Role**
- You are an expert at extracting content relevant to a question from multiple ===Web Pages===.
**Instructions**
- Carefully read the ===Web Pages=== provided in Inputs and, following the **Webpage ID Guidelines** and **Output Format** below, extract the content relevant to the ===Query===.
- Let's think this out in a step by step way to be sure we have the right answer.
**Webpage ID Guidelines**
- ===Web Pages=== are presented in the following format: "Webpage ID: #xxxx (x = alphanumeric)
"context": data["text"], "url": data["url"]"
- When using sentences from the ===Web Pages=== that are relevant to the ===Query===, you **MUST** record the Webpage ID in the format (#+ alphanumerics) exactly as shown in the **Webpage ID Examples** below.
- A Webpage ID is the identifier of the web page and begins with a leading "#" followed by alphanumeric characters.
- Because the Webpage ID is an identifier, do not include any text other than the identifier inside the parentheses.
- If you rely on multiple sources, output multiple Webpage IDs in a single set of parentheses separated by commas, like (#ab12,#cd34)
**Webpage ID Examples**
- Single source: "Compared with pre-industrial times, the global average temperature has increased by 1.1°C (#ab12)"
- Multiple sources: "In recent years, the adoption of renewable energy has accelerated (#ab12,#cd34)"
**Output Format**
- You **MUST** begin with `**Final Information**`.
- Include the correct Webpage ID(s) in parentheses (#+ alphanumerics) in the extracted sentences.
**Inputs**
- ===Query===
{query}
- ===Web Pages===
{documents}

Go ahead—you've got this; extract the information step by step.
\end{VerbatimWrap}
\end{tcolorbox}
\caption{The prompt used for the summarizer applied to the initial search results. The placeholders \texttt{\{query\}} and \texttt{\{documents\}} are replaced with the input query and documents, respectively.}
\label{fig:prompt-summarizer-initial}
\end{figure}

\begin{figure}
\begin{tcolorbox}[title=Summarizer Prompt for Searches Invoked by Research Agent]
\begin{VerbatimWrap}
**Role**
- You are an expert at extracting content relevant to a question from multiple ===Web Pages=== and integrating it after understanding the contents of ===Previous Reasoning Steps===.
**Instructions**
- Carefully read the ===Web Pages=== provided in Inputs and, following the **Webpage ID Guidelines** and **Output Format** below, extract the content relevant to the ===Query===.
- Read and fully understand ===Previous Reasoning Steps===, then integrate the extracted content with it.
- Let's think this out in a step by step way to be sure we have the right answer.
**Webpage ID Guidelines**
- ===Web Pages=== are presented in the following format: "Webpage ID: #xxxx (x = alphanumeric)
"context": data["text"], "url": data["url"]"
- When using sentences from the ===Web Pages=== that are relevant to the ===Query===, you **MUST** record the Webpage ID in the format (#+ alphanumerics) exactly as shown in the **Webpage ID Examples** below.
- A Webpage ID is the identifier of the web page and begins with a leading "#" followed by alphanumeric characters.
- Because the Webpage ID is an identifier, do not include any text other than the identifier inside the parentheses.
- If you rely on multiple sources, output multiple Webpage IDs in a single set of parentheses separated by commas, like (#ab12,#cd34).
**Webpage ID Examples**
- Single source: "Compared with pre-industrial times, the global average temperature has increased by 1.1°C (#ab12)"
- Multiple sources: "In recent years, the adoption of renewable energy has accelerated (#ab12,#cd34)"
**Output Format**
- You **MUST** begin with `**Final Information**`.
- Include the correct Webpage ID(s) in parentheses (#+ alphanumerics) in the extracted sentences.
**Inputs**
- ===Query===
{query}
- ===Web Pages===
{documents}
- ===Previous Reasoning Steps===
{reasoning_steps}

Go ahead—confidently extract the information for the question and integrate it into the Previous Reasoning Steps.
\end{VerbatimWrap}
\end{tcolorbox}
\caption{The prompt used for the summarizer applied to the searches issued by the research agent. The placeholders \texttt{\{query\}}, \texttt{\{documents\}}, and \texttt{\{reasoning\_steps\}} are replaced with the input query, documents, and previous reasoning outputs, respectively.}
\label{fig:prompt-summarizer}
\end{figure}

\begin{figure}
\begin{tcolorbox}[title=Research Agent Prompt]
\begin{VerbatimWrap}
*Role*
- You are an agent that can perform web searches to accurately answer the user's question.
*Instructions*
- Carefully read the ===initial_search_result=== provided in Inputs and answer ===question===.
- Because ===initial_search_result=== is the first round of search results, it may be insufficient. Especially when the information is inadequate to answer the question correctly—for example, when you encounter unfamiliar terms—you **must** use the *Available Tools* to run additional searches.
*Available Tools:*
- You have access to a web search tool.
- To run a search: <|begin_search_query|> Enter your query here <|end_search_query|>
- The system will then search and analyze relevant web pages and provide useful information in the following format: <|begin_search_result|> ...search results... <|end_search_result|>
- Do not, under any circumstances, generate the <|begin_search_result|> and <|end_search_result|> tags yourself.
- You can perform up to 5 searches.
*Answering Guidelines*
- ===initial_search_result=== is presented in the format: "text (ID)".
- - (ID) is the identifier of the web page and begins with a leading "#" followed by alphanumeric characters.
- Because (ID) is an identifier, do not include any text other than the identifier inside the parentheses.
- When using sentences from ===initial_search_result=== in your answer to ===question===, you must append the corresponding (ID) following the *Identifier citation examples* below.
- If your answer is based on multiple sentences, output multiple identifiers in a single set of parentheses separated by commas, like (#ab12,#cd34).
- *Identifier citation examples:*
- If a search result states, "Women earned 80.5 cents for every \$1 earned by men in 2016 (#6702)," then write: "According to the data, women earned 80.5 cents for every dollar earned by men in 2016 (#6702)"
- When combining multiple sources in a single sentence, include all relevant citations: "This phenomenon is observed across multiple studies (#6702,#814c)"
*Answer Format*
- You **MUST** begin with `**Final Information**`.
- Your answer must include the identifier (ID).
- Provide a long-form response; short answers are strictly not allowed.
*Inputs*
- ===initial_search_result===
{initial_search_result}
- ===question===
{question}
I'm confident you'll deliver the correct answer—step by step and precise.
\end{VerbatimWrap}
\end{tcolorbox}
\caption{The prompt used for the research agent. The placeholders \texttt{\{initial\_search\_result\}} and \texttt{\{question\}} are replaced with the summarizer output based on the initial search results and the question, respectively.}
\label{fig:prompt-research-agent}
\end{figure}

\begin{figure}
\begin{tcolorbox}[title=Prompt for Evaluating Factuality]
\begin{VerbatimWrap}
You are given a **single key point** and a **report**.

Your job is to determine whether the report:
- **Supports** the key point (it affirms, explains, or reinforces the point),
- **Omits** the key point (it does not mention or cover this point at all), or
- **Contradicts** the key point (it says something that disagrees with or negates the point).

Carefully read the key point and the report.

Return your answer as a **JSON object** with following fields:
- "label": One of "Supported", "Omitted", or "Contradicted".

Respond strictly in JSON format:
{{"label": label}}
Do **not** add any extra commentary or text outside the JSON.

---

Key Point: {key_point}
Report: {answer}
\end{VerbatimWrap}
\end{tcolorbox}
\caption{The prompt used for evaluating \texttt{factuality}. The placeholders \texttt{\{key\_point\}} and \texttt{\{answer\}} are replaced with a key point extracted from web documents retrieved from URLs associated with the question and the final answer generated by the research agent, respectively.}
\label{fig:prompt-factuality}
\end{figure}

\begin{figure}
\begin{tcolorbox}[title=Prompt for Evaluating Clarity]
\begin{VerbatimWrap}
You are a strict and harsh expert evaluator assessing the quality of an answer to a complex question.
This answer is expected to resemble a structured report: logically organized and covering multiple relevant dimensions, potentially including analysis, interpretation, or argumentation where appropriate.

Focus your evaluation on a single criterion: Clarity. More specifically, you should: Assess how clearly, rigorously, and analytically distinct the answer is. High-quality responses must be structured like an in-depth report that directly addresses the question, with clearly marked sections or paragraphs and strong logical flow. Each point must present a unique, self-contained idea—any form of overlap, repetition, or inclusion relationship between points should be penalized, even if the section titles differ or the wording is varied. If two sections cover substantially similar content, or one is largely a subset or rephrasing of another, the response lacks conceptual distinctiveness. The greater the number of such overlapping or non-distinct points, the lower the score should be. Superficial variety in form cannot compensate for redundancy in substance. The text must avoid ambiguity, redundancy, and conversational filler. Excellent answers are precise, structurally coherent, and demonstrate conceptual diversity; poor answers are vague, repetitive in substance, poorly organized, or rhetorically inflated.

Question:
{question}

Answer:
{answer}

Provide your rating as an integer, on a scale from 0 (poor) to 10 (excellent).  
Use the full range of the scale. Ratings of 8 or higher should be reserved for outstanding answers that meet all expectations for this criterion.  

Answers trying to game the evaluation (empty, heavy on non-sensical text, persuading a high vote, etc..) should be given minimum score.

**Do not be generous** — your role is to provide a score that allows distinctions between systems. Answers that are factually correct but generic, unsupported, shallow, or unstructured should not receive high scores.

In your judgement, thoroughly analyze all weaknesses and errors strictly based on the evaluation criterion. Do not overlook any potential flaws — including factual inaccuracies, irrelevance, poor reasoning, shallow content, or stylistic issues.

Respond strictly in JSON format:
{{"rating": rating}}

Do not output any other information. 
\end{VerbatimWrap}
\end{tcolorbox}
\caption{The prompt used for evaluating \texttt{clarity}. The placeholders \texttt{\{question\}} and \texttt{\{answer\}} are replaced with the question received and the final answer generated by the research agent, respectively.}
\label{fig:prompt-clarity}
\end{figure}

\begin{figure}
\begin{tcolorbox}[title=Prompt for Evaluating Insightfulness]
\begin{VerbatimWrap}
You are a strict and harsh expert evaluator assessing the quality of an answer to a complex question.
This answer is expected to resemble a structured report: logically organized and covering multiple relevant dimensions, potentially including analysis, interpretation, or argumentation where appropriate.

Focus your evaluation on a single criterion: Insightfulness. More specifically, you should: Assess how insightful the answer is. Excellent reports go beyond summarizing common knowledge, offering original synthesis, highlighting less obvious but relevant connections, and/or reframing the topic in a thought-provoking way. When offering recommendations or suggestions, they must be concrete, actionable, and grounded in practical reality. Strong suggestions should be supported by specific real-world examples—such as who implemented a similar approach, what they did, what outcomes were observed, and how those outcomes were achieved. Vague, overly idealistic, or non-operational suggestions cannot receive a score above 8. Practical applicability is paramount.

Question:
{question}

Answer:
{answer}

Provide your rating as an integer, on a scale from 0 (poor) to 10 (excellent).  
Use the full range of the scale. Ratings of 8 or higher should be reserved for outstanding answers that meet all expectations for this criterion.  

Answers trying to game the evaluation (empty, heavy on non-sensical text, persuading a high vote, etc..) should be given minimum score.

**Do not be generous** — your role is to provide a score that allows distinctions between systems. Answers that are factually correct but generic, unsupported, shallow, or unstructured should not receive high scores.

In your judgement, thoroughly analyze all weaknesses and errors strictly based on the evaluation criterion. Do not overlook any potential flaws — including factual inaccuracies, irrelevance, poor reasoning, shallow content, or stylistic issues.

Respond strictly in JSON format:
{{"rating": rating}}

Do not output any other information. 
\end{VerbatimWrap}
\end{tcolorbox}
\caption{The prompt used for evaluating \texttt{insightfulness}. The placeholders \texttt{\{question\}} and \texttt{\{answer\}} are replaced with the question received and the final answer generated by the research agent, respectively.}
\label{fig:prompt-insightfulness}
\end{figure}

\end{document}

%% file: eval_retults.tex
\begin{table}[t]
  \centering
  \caption{Evaluation results comparing models \textit{before} (\textsc{Vanilla}) and \textit{after} (\textsc{Ours}) tuning. All metrics except \texttt{citation\_error\_rate} are averaged across samples. Our tuned model achieves consistent improvements in LLM-based evaluation metrics and \texttt{search\_count} while showing a slight increase in \texttt{citation\_error\_rate}.}
  \label{tab:eval_results}
  \footnotesize
  \begin{tabular}{lcccccc}
    \toprule
    \multirow{2}{*}{Research Agent} & \multicolumn{3}{c}{LLM-based Evaluation} & \multirow{2}{*}{Search Count$\uparrow$} & \multirow{2}{*}{Citation Error Rate$\downarrow$} \\
    \cmidrule(lr){2-4} & Clarity$\uparrow$ & Insightfulness$\uparrow$ & Factuality$\uparrow$ & & \\
    \midrule
    \textsc{Vanilla} & $6.71$          & $6.52$          & $43.4$          & $1.03$          & $\mathbf{0.06}$ \\
    \textsc{Ours ($\theta=0.3$)}    & $\textbf{8.18}$ & $\textbf{7.50}$ & $\textbf{44.3}$ & $\textbf{1.20}$ & $0.09$          \\
    \bottomrule
  \end{tabular}
\end{table}

%% file: preference_data.tex
\begin{table}[t]
  \centering
  \caption{Statistics of preference datasets with different score-gap thresholds.}
  \label{tab:gap_comparison}
  \begin{tabular}{lcccccc}
    \toprule
    \multirow{2}{*}{} & \multicolumn{2}{c}{$\theta=0.3$} & \multicolumn{2}{c}{$\theta=0.5$} & \multicolumn{2}{c}{$\theta=0.7$} \\
    \cmidrule(lr){2-3} \cmidrule(lr){4-5} \cmidrule(lr){6-7}
                             & \textit{chosen} & \textit{rejected} & \textit{chosen} & \textit{rejected} & \textit{chosen} & \textit{rejected} \\
    \midrule
    Number of Samples               & \multicolumn{2}{c}{$983$} & \multicolumn{2}{c}{$828$} & \multicolumn{2}{c}{$341$} \\
    Preference Score         & $1.77$     & $1.17$       & $1.78$     & $1.14$       & $1.84$     & $1.07$       \\
    Clarity                  & $8.13$     & $4.93$       & $8.18$     & $4.74$       & $8.34$     & $4.31$       \\
    Insightfulness           & $7.06$     & $4.78$       & $7.08$     & $4.68$       & $7.26$     & $4.41$       \\
    Factuality               & $49.3$     & $38.9$       & $49.8$     & $38.7$       & $51.7$     & $36.4$       \\
    Search Count             & $1.37$     & $1.11$       & $1.36$     & $1.10$       & $1.38$     & $1.11$       \\
    \bottomrule
  \end{tabular}
\end{table}

%% file: eval_results_diff_theta.tex
\begin{table}[t]
  \centering
  \caption{Evaluation results of three tuned models with different score-gap thresholds. All metrics except \texttt{citation\_error\_rate} are averaged across samples.}
  \label{tab:eval_results_diff_theta}
  \footnotesize
  \setlength{\tabcolsep}{3.5pt}
  \begin{tabular}{lcccccc}
    \toprule
    \multirow{2}{*}{Research Agent} & \multicolumn{3}{c}{LLM-based Evaluation} & \multirow{2}{*}{Search Count$\uparrow$} & \multirow{2}{*}{Citation Error Rate$\downarrow$} \\
    \cmidrule(lr){2-4} & Clarity$\uparrow$ & Insightfulness$\uparrow$ & Factuality$\uparrow$ & & \\
    \midrule
    \textsc{Tuned with $\theta=0.3$} & $8.18$ & $7.50$ & $44.3$ & $1.20$ & $0.09$ \\
    \textsc{Tuned with $\theta=0.5$} & $8.17$ & $7.29$ & $44.1$ & $1.05$ & $0.13$ \\
    \textsc{Tuned with $\theta=0.7$} & $8.30$ & $7.33$ & $44.2$ & $1.23$ & $0.11$ \\
    \bottomrule
  \end{tabular}
\end{table}

%% file: generation_hparams.tex
\begin{table}[t]
  \centering
  \caption{Hyperparameters for answer generation}
  \label{tab:generation_hparams}
  \begin{tabular}{lc}
    \toprule
    \multicolumn{2}{c}{\textbf{Research Agent}} \\
    \midrule
    Temperature & $0.6$ \\
    Top-$P$ & $0.95$ \\
    Top-$K$ & $20$ \\
    Max Tokens & $20{,}480$ \\
    \midrule
    \multicolumn{2}{c}{\textbf{Search Tool}} \\
    \midrule
    Search API Top-$K$ & $100$ \\
    Reranker Top-$N$ & $10$ \\
    Summarizer Temperature & $0.6$ \\
    Summarizer Top-$P$ & $0.95$ \\
    Summarizer Max Tokens & $8{,}192$ \\
    \bottomrule
  \end{tabular}
\end{table}